\documentclass[10pt,twocolumn,numbers]{article}

\usepackage{graphicx}
\usepackage{amssymb}
\usepackage{amsthm,amsmath,bbm,dsfont}
\usepackage{mathtools}

\usepackage{lineno}
\usepackage{caption}

\usepackage[colorlinks=true, citecolor=blue]{hyperref}
\usepackage[linesnumbered,lined,ruled,commentsnumbered]{algorithm2e}
\usepackage{algcompatible}

\usepackage{wrapfig}
\usepackage[numbers,sort&compress]{natbib}
\usepackage{comment}
\usepackage{longtable}
\usepackage{multirow}
\usepackage{bbding}
  
\usepackage[bottom]{footmisc}
\usepackage{booktabs}
\usepackage{pifont}
\usepackage{enumitem}
\setlist[itemize]{noitemsep, topsep=0pt, leftmargin=10pt}
\setlist[enumerate]{noitemsep, topsep=0pt}
\setlist[description]{noitemsep, topsep=0pt}
\usepackage{float}
\usepackage{subfigure}
\usepackage{wrapfig}
\usepackage{framed,xcolor}
\usepackage{newfloat}
\usepackage[xcolor,framemethod=TikZ]{mdframed}
\usepackage{amsthm}
\usepackage{thmtools}
\usepackage{authblk}
\patchcmd{\thebibliography}{\section*{\refname}}{}{}{}
\usepackage{tikz}
\usetikzlibrary{arrows,shapes,chains,calc,patterns,decorations.pathreplacing}
\usepackage{stmaryrd}
\usepackage{fancyhdr}
\usepackage{subfigure}
\usepackage{comment}
\def\BibTeX{{\rm B\kern-.05em{\sc i\kern-.025em b}\kern-.08em
	T\kern-.1667em\lower.7ex\hbox{E}\kern-.125emX}}

\newtheoremstyle{theoremdd}
{\topsep}
{\topsep}
{\itshape}
{0pt}
{\fontfamily{cmss}\selectfont\bfseries}
{.}
{ }
{\thmname{#1}\thmnumber{ #2}\thmnote{ (#3)}}

\theoremstyle{theoremdd}

\mdfdefinestyle{myStyle}{roundcorner=5pt,backgroundcolor=brown!20,linecolor=red!40!black,linewidth=1.5pt}

\usepackage{titlesec}
\titleformat*{\section}{\fontfamily{cmss}\selectfont\large\bfseries}
\titleformat*{\subsection}{\fontfamily{cmss}\selectfont\normalsize\bfseries}
\titleformat*{\subsubsection}{\fontfamily{cmss}\selectfont\normalsize}
\usepackage[left=0.7 in, right=0.7 in, top=1.0 in, bottom=1.0 in]{geometry}
\graphicspath{{./Figures/}}

\renewcommand\abstractname{\fontfamily{cmss}\selectfont\normalsize\bfseries\textbf{Abstract}}

\renewenvironment{abstract}{%
\centering\small
\list{}{\leftmargin1.5cm \rightmargin\leftmargin}
\item\relax

\begin{mdframed}[]
\item[\hskip\labelsep\scshape\abstractname.]%
}{%
\end{mdframed}
\endlist \par\bigskip
}
\makeatletter
\patchcmd{\@maketitle}{\LARGE \@title}{\fontfamily{cmss}\selectfont\LARGE\@title}{}{}
\makeatother

\graphicspath{{./figures/}}

\begin{document}

\title{Learning-based solutions to nonlinear hyperbolic PDEs: Empirical insights on generalization errors}

%


\author[1,2]{Bilal Thonnam Thodi}
\author[2]{Sai Venkata Ramana Ambadipudi}
\author[1,2]{Saif Eddin Jabari}

\affil[1]{New York University Tandon School of Engineering, Brooklyn NY, U.S.A.}
\affil[2]{New York University Abu Dhabi, Saadiyat Island, P.O. Box 129188, Abu Dhabi, U.A.E.}

\date{}



\twocolumn[
\begin{@twocolumnfalse}
	
	\maketitle

\begin{abstract}
    We study learning weak solutions to nonlinear hyperbolic partial differential equations (H-PDE), which have been  difficult to learn due to discontinuities in their solutions. We use a physics-informed variant of the Fourier Neural Operator ($\pi$-FNO) to learn the weak solutions. We empirically quantify the generalization/out-of-sample error of the $\pi$-FNO solver as a function of input complexity, i.e., the distributions of initial and boundary conditions. 
    Our testing results show that $\pi$-FNO generalizes well to unseen initial and boundary conditions. We find that the generalization error grows linearly with input complexity. Further, adding a physics-informed regularizer improved the prediction of discontinuities in the solution. We use the  Lighthill-Witham-Richards (LWR) traffic flow model as a guiding example to illustrate the results.

    
\end{abstract}
\bigskip
\end{@twocolumnfalse}
]

\section{Introduction}

Hyperbolic partial differential equations (H-PDEs) arise in the study of nonlinear wave motion in applications such as the behavior of water waves, vehicular traffic flow, and even in high-speed physics like blast waves and sonic booms. Typically, (inviscid) H-PDEs are of the form $u_t + c u_x = 0$, where $u$ is a conserved quantity. Wave motion in H-PDEs is characterized by finite propagation speeds $c$. The solutions are characteristic trajectories emanating from the datum (e.g., the initial conditions). 
Nonlinear H-PDEs, where the nonlinearity results in different wave propagation speeds, e.g., $c = c(u)$, result in characteristic lines that may cross somewhere in the domain, and at those points, the solution is multi-valued, which means that the problem does not have a solution in the classical sense. Here, weak solutions that permit discontinuities are introduced, which are conventionally solved using finite volume or finite difference-based numerical schemes \cite{leveque1992numerical,witham1974waves}. 

Recently, deep learning-based (DL) methods have appeared that aim to overcome the limitations of conventional numerical solvers, namely, high computational cost, grid dependence, and knowledge of complete initial and boundary conditions. The DL solvers have shown remarkable results for both forward and inverse problems, especially where the PDE solutions are smooth \cite{raissi2019pinns,li2021fno}. However, DL solutions having high irregularities (e.g., discontinuities), such as those arising in the weak solutions of H-PDEs, are only partially successful \cite{torrado2022pinns_buckly,zhang2022pinns_implicit,patel2022cvpinns,jagtap2020cpinns}. This is partly due to the ill-posedness of H-PDEs, i.e., derivatives are not defined everywhere, and the H-PDE residuals do not form a correct physics loss metric. Further, these studies lacked a systematic procedure to evaluate the out-of-sample performance or the generalization error. 

To this end, we explore Fourier Neural Operators (FNO) \cite{li2021fno} for learning weak solutions of H-PDEs. We propose a systematic training and testing experiment where the FNO solver is trained with solutions of elementary input conditions and evaluated for solutions of general input conditions. We quantify the empirical generalization error of the FNO solver as a function of the input complexity, i.e., how the out-of-sample error grows as the distribution of input conditions becomes more general. To capture physically consistent weak solutions (e.g., shocks), we propose using an integral form of the H-PDE (written in discrete form) as the physics loss function instead of the H-PDE residual. We demonstrate these results using vehicular traffic flow as an example.


\section{Methods}

\paragraph{Problem setting} 

We consider the LWR traffic flow model \cite{light1955lwr,richards1956lwr} as the guiding example for nonlinear scalar H-PDEs. The LWR model is a continuum description of the flow of vehicles on a road. Consider a space-time domain $\Omega \subset \mathbb{R} \times \mathbb{R}_{+}$. Denote by $u(x,t) : \Omega \rightarrow [0, u_{\max}]$ the density of traffic at position $x \in \mathbb{R}$ and time $t \in \mathbb{R}_+$. Here density refers to the average number of vehicles per unit length. Let $q(x,t) : \Omega \rightarrow [0, q_{\max}]$ the traffic flux, which is the number of vehicles crossing a point per unit time. The LWR model describes the evolution of traffic density based on the principle of vehicular conservation:
\begin{equation} \label{eqn1:lwr0}
    u_t + q_x = 0; ~~
    u(x,0) = \bar{u}_0; ~~
    u(x_b,t) = \bar{u}_b; ~~
    (x,t) \in \Omega; ~~
\end{equation}
where $u_t \equiv \frac{\partial u}{\partial t}$, $\bar{u}_0$ is an initial condition, and $\bar{u}_b$ is a boundary condition. For the LWR model, one typically prescribes a flux function $q = f(u)$, where $f(u): [0, u_{\max}] \rightarrow [0, q_{\max}]$, which is concave in the context of traffic flow. The LWR model is given as
\begin{equation} \label{eqn1:lwr}
    u_t + f'(u) u_x = 0; ~~
    u(x,0) = \bar{u}_0; ~~
    u(x_b,t) = \bar{u}_b; ~~
    (x,t) \in \Omega; ~~
\end{equation}
We refer to \eqref{eqn1:lwr} as the \emph{forward} problem if $\bar{u}_b$ corresponds to solutions at the boundary points $(x_b,t) \in \partial \Omega$ and as the \emph{inverse} problem if $\bar{u}_b$ is replaced with densities at random points in the domain $\Omega$ i.e., $(x_b,t) \in \Omega$. 
We use $f(u) = uv_{\rm max} (1 - u/u_{\rm max})$, where $u_{\rm max}$ is the maximum traffic density and $v_{\rm max}$ is the maximum traffic speed.
A major difficulty in solving the \emph{forward} problem is handling discontinuities when they appear in the solution due to nonlinearity in the system.
The \emph{inverse} problem poses the additional challenge that the boundary condition is unknown, and the solution needs to be inferred from partially observed measurements. 
We tackle both challenges using a deep learning-based solver, discussed below.

\paragraph{Fourier Neural Operator solver}

We are interested in learning the solution operator that maps the input function $a := (\bar{u}_0, \bar{u}_b)$ to the weak solution $u(x,t)$ over $\Omega$. Let the input function be $a \in \mathcal{A}$ and output function be $u \in \mathcal{U}$. The problem \eqref{eqn1:lwr} can be rephrased as one of learning an operator $\mathcal{G}: \mathcal{A} \rightarrow \mathcal{U}$. We approximate $\mathcal{G}$ using the Fourier Neural Operator (FNO) of \citet{li2021fno}, represented by the parametric model $\mathcal{G}_{\Theta}$, and given by
\begin{equation} \label{eqn2:fno}
    \widehat{u} = \mathcal{G}_{\Theta} (a) = \big( \mathcal{Q} \circ \mathcal{F}^{(L)} \circ \mathcal{F}^{(L-1)} \circ \cdots \circ \mathcal{F}^{(2)} \circ \mathcal{F}^{(1)} \circ \mathcal{P} \big) (a)
\end{equation}
where $\{ \mathcal{F}^{(l)} \}_{l=1}^{L}$ is a set of Fourier operators while $\mathcal{P}$ and $\mathcal{Q}$ are projection operators. A single Fourier operator is defined as
\begin{equation} \label{eqn3:fourier}
    \mathcal{F} (z) = \sigma \Big( W \cdot z + \mathrm{IFFT} \big( R \cdot \mathrm{FFT} ( z ) \big) \Big)
\end{equation}
for any latent input $z$, where FFT and IFFT denote the Fourier transform and its inverse. $\Theta = \{W^{(l)}, R^{(l)}\}_{l=1}^{L}$ is the set of trainable parameters of the FNO operator $\mathcal{G}_{\Theta}$. Our motivation for using the FNO operator \eqref{eqn2:fno} is its efficient approximation in the Fourier domain and it ability to learn complex dynamics \cite{li2021fno}. The parameter complexity involved in learning $\mathcal{G}_{\Theta}$ depends on the size of $R$, which is independent of the domain size $|\Omega|$.

\paragraph{Physics-informed training}

The FNO model \eqref{eqn2:fno} can be trained end-to-end in a supervised learning framework over an appropriately defined loss function. We perform physics-informed training where the loss function has two parts $-$ an empirical training data loss $L_{\rm data}$ and a physics constraint loss $L_{\rm phys}$ to emulate the PDE operator. $L_{\rm data}$ is simply $\sum_{n \in N} || \mathcal{G}_{\Theta}(a^{(n)}) - u^{(n)} ||_2$, where $N$ is the number of samples and  $u^{(n)}$ is the $n^{th}$ sample solution. 

One could use the PDE residual of \eqref{eqn1:lwr} to form the physics loss $L_{\rm phys}$, as in the conventional physics-informed neural networks \cite{raissi2019pinns}. However, \eqref{eqn1:lwr} is not a well-posed PDE, i.e., derivatives are not defined everywhere, especially near discontinuities and hence not a well-defined loss metric. Thus, we resort to the integral form of \eqref{eqn1:lwr} to form the physics loss $L_{\rm phys}$ as follows:
\begin{multline} \label{eqn4:loss_phys}
    L_{\rm phys} = \Big\| \big\langle u(x,t+\Delta t) - u(x, t) \\+ \frac{\Delta t}{\Delta x} \left[ q(x-\Delta x/2, t) - q(x+\Delta x/2, t) \right] \big\rangle_{(x,t)} \Big\|_2,
\end{multline}
where $u(x,t)$ and $q(x,t)$ are defined over a discretization of $\Omega$, and $\langle \cdot \rangle_{(x,t)}$ is a concatenation operator. Accordingly, we train two different FNO models with two different objective functions, as shown below:
\begin{align}
        &\text{(i) FNO model :} ~\underset{\Theta}{\min} ~ L_{\rm data} \\
        &\text{(ii) $\pi$-FNO model :} ~\underset{\Theta}{\min} ~ L_{\rm data} + \lambda L_{\rm phys}
\end{align}

\paragraph{Data and training experiments} We obtain the training and testing dataset by numerically simulating \eqref{eqn1:lwr} for different input conditions. The numerical scheme used for generating the datasets and the FNO training codes is described in Appendix \ref{app:godunov} and Appendix \ref{app:data}. 

\begin{figure*}[hbt!]
  \centering
  \includegraphics[width=0.80\textwidth]{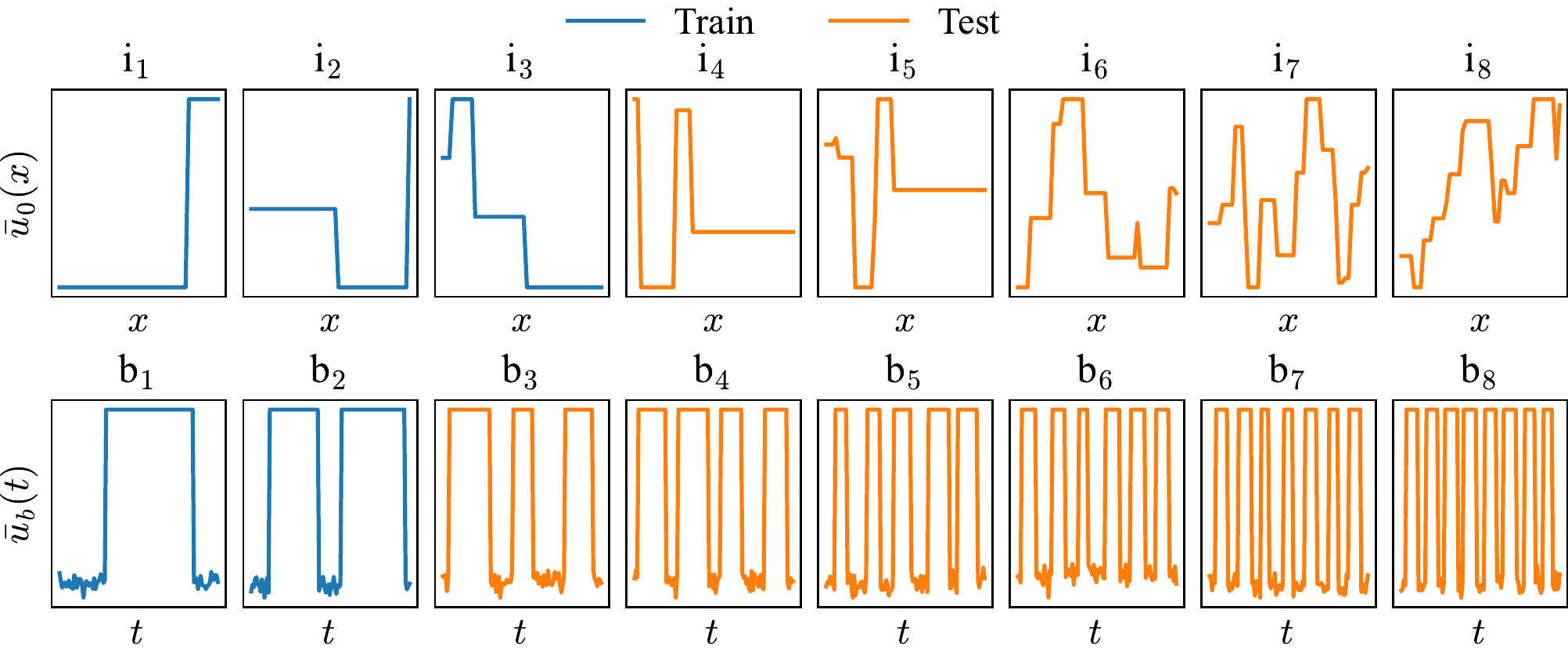}
  \caption{Initial and boundary conditions used for training and testing the FNO solver.}
  \label{fig:inp_conds}
\end{figure*}

We perform systematic training and testing experiments to quantify the generalization performance of the FNO solver. During training, the FNO solver is shown solutions of simple dynamics, for instance, generated from step-wise initial conditions (a single vehicle queue) and one or two stepped wavelet-like boundary conditions (emulating vehicles stopping at traffic lights). The FNO solver is then tested with solutions of complex dynamics generated from general initial conditions (multiple vehicle queues) and multi-stepped wavelet-like boundary conditions (multiple stops at traffic lights). The set of initial and boundary conditions used in training and testing are shown in Figure \ref{fig:inp_conds}. Training data consists of solutions using inputs (${\rm i}_0$-${\rm i}_3$, ${\rm b}_0$-${\rm b}_2$). Testing data are made of (${\rm i}_0$-${\rm i}_9$, ${\rm b}_0$-${\rm b}_2$) for evaluating initial conditions, and (${\rm i}_0$-${\rm i}_3$, ${\rm b}_0$-${\rm b}_8$) for evaluating boundary conditions.

The goal is to train the FNO solver with simple solutions and assess the out-of-sample error as input conditions become complex. For the LWR traffic flow example, the input complexity refers to (a) how the vehicles are distributed spatially at time $t=0$ (i.e., $\bar{u}_0$) and (b) the impact of traffic signals at the road exit $x=x_{\rm max}$ (i.e., boundary condition $\bar{u}_b$). These two input factors put together can generate complex dynamics $u$. 
Training details are summarized in Appendix \ref{app:hyper}.

\section{Results}

\begin{figure}[hbt!]
  \centering
  \includegraphics[width=0.45\textwidth]{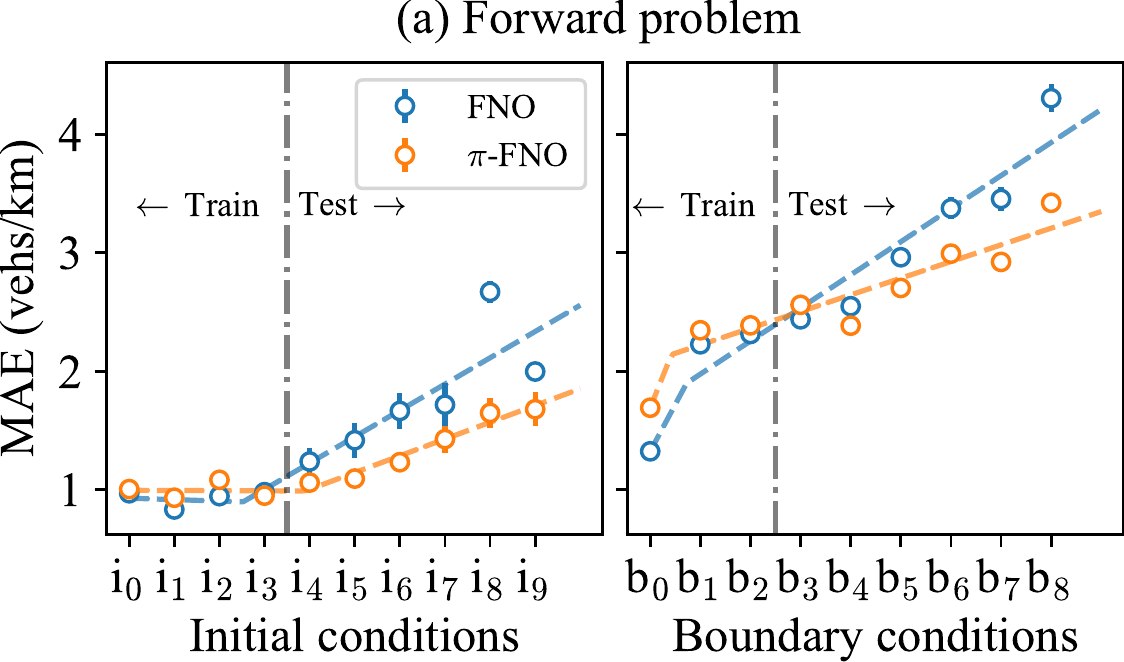} \hspace{0.04in} 
  \includegraphics[width=0.45\textwidth]{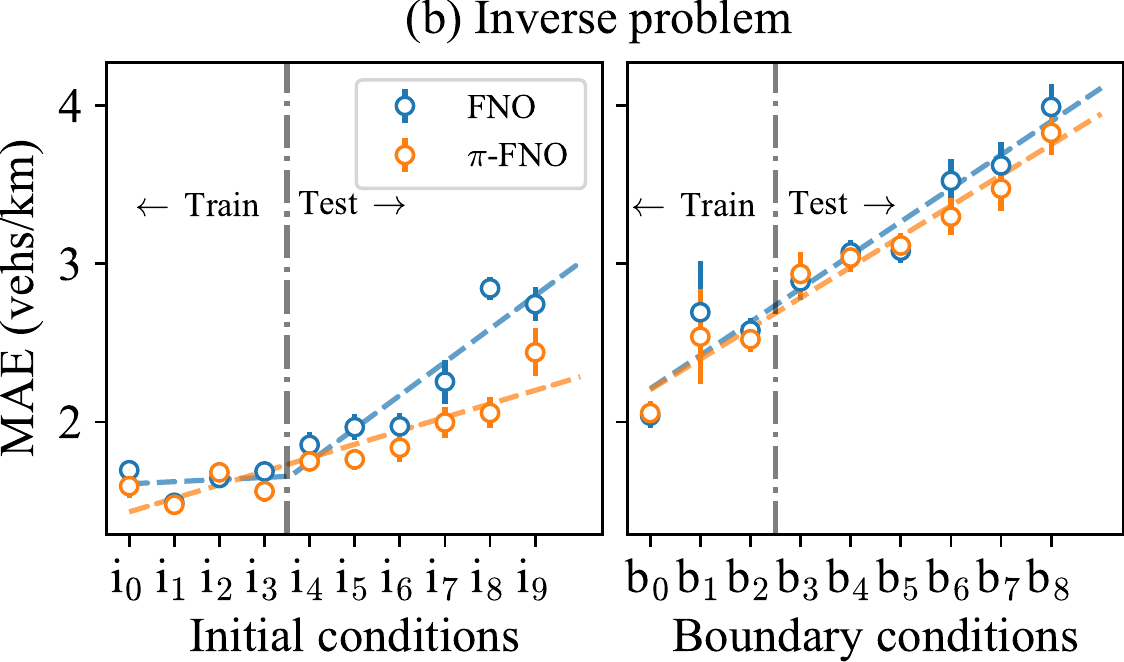}
  \caption{Empirical generalization (out-of-sample) error as function of input conditions.}
  \label{fig:outofsample}
\end{figure}

\paragraph{Generalization error v/s input complexity} The out-of-sample errors as a function of input conditions are summarized in Figure \ref{fig:outofsample}. Each data point is the average mean absolute error (MAE) of $50$ samples. A piece-wise linear trendline is fitted to the error plots. The trendline shows that MAE is nearly constant for input conditions seen during training (i$_0$-i$_3$, b$_0$-b$_2$) and steadily increases for input conditions at testing (i$_4$-i$_9$, b$_3$-b$_9$).

     We define the generalization error as the slope of this trendline, summarized in Table 1. For instance, Figure \ref{fig:outofsample} says that MAE increases by $0.141$ vehs/km ($+0.12\%$) for every additional traffic light at the road exit $\bar{u}_b$ using $\pi$-FNO model. Similarly, an additional non-uniformity in the initial condition $\bar{u}_0$ increases MAE by $0.142$ vehs/km ($+0.12\%$). 
    \begin{table}
   \centering
    \label{tab:error_rates}
    \captionof{table}{\small Error rates (vehs/km)}
    \begin{tabular}{@{}lcccc@{}}
    \toprule
     & \multicolumn{2}{c}{FNO} & \multicolumn{2}{c}{$\pi$-FNO} \\ \cmidrule(l){2-5} 
     & $\bar{u}_0$ & $\bar{u}_b$ & $\bar{u}_0$ & $\bar{u}_b$ \\ \cmidrule(l){2-5} 
    Forward & 0.222 & 0.279 & 0.142 & 0.141 \\
    Inverse & 0.210 & 0.211 & 0.085 & 0.194 \\ \bottomrule
    \end{tabular}
    \end{table}

Figure \ref{fig:outofsample} concludes that the generalization error for the testing set grows linearly with the input complexity for both the forward and inverse problems. Also, $\pi$-FNO incurs lower error rates compared to the FNO model.

\paragraph{Sample predictions}

\begin{figure*}[hbt!]
  \centering
  \includegraphics[width=.90\textwidth]{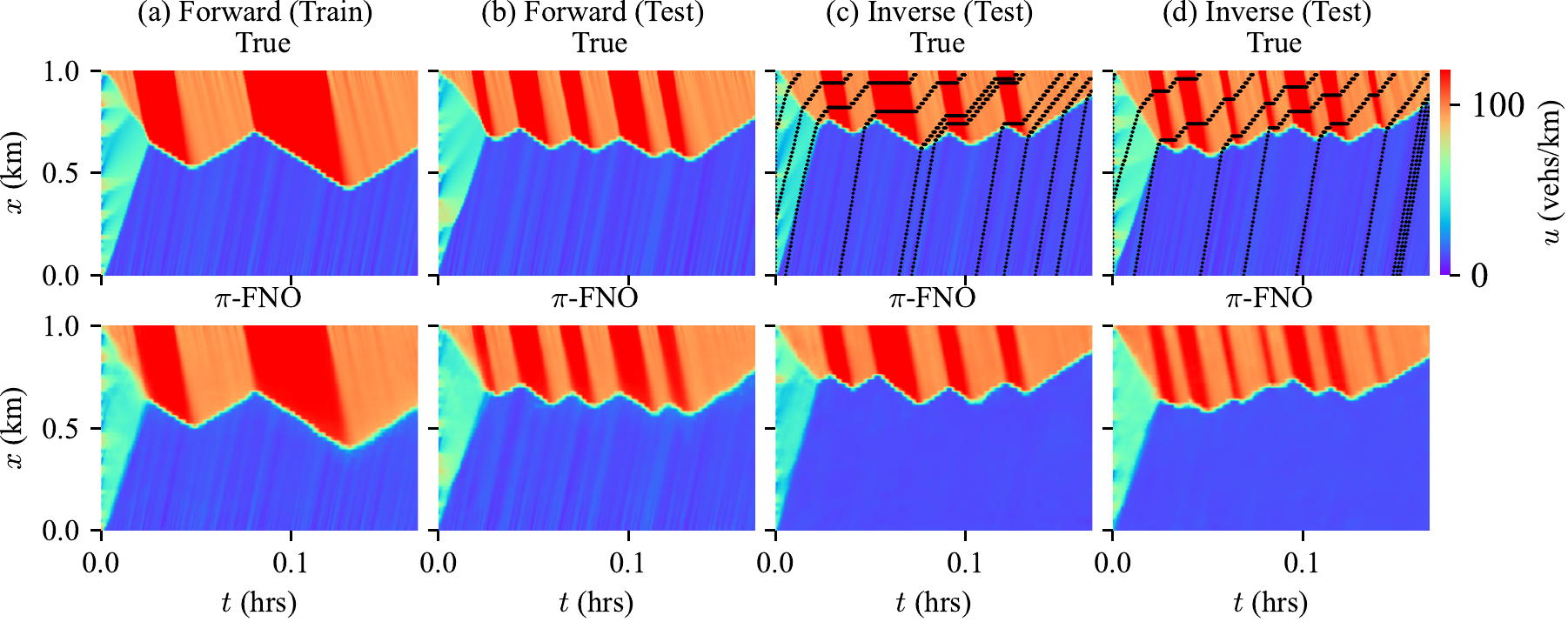}
  \caption{Comparison of true and $\pi$-FNO predicted solutions. Sub-figure (a) is a training scenario, and (b)-(d) are test scenarios. The dotted curve is the input locations of $\bar{u}_b$ for the inverse problem.}
  \label{fig:sample_preds}
\end{figure*}

The predicted and true solutions for four different input conditions are shown in Figure \ref{fig:sample_preds}. 

The $\pi$-FNO solver is only shown density dynamics similar to Figure \ref{fig:sample_preds}a at the training stage. Figure \ref{fig:sample_preds}b-\ref{fig:sample_preds}d are the density dynamics that $\pi$-FNO solver generalizes flawlessly. This implies that the $\pi$-FNO solver learned to capture the traffic queuing dynamics (i.e., vehicle queue formation and dissipation) as a function of the boundary flows. Also, the inverse problem results in Figure \ref{fig:sample_preds}c, and \ref{fig:sample_preds}d shows that $\pi$-FNO can qualitatively recover solution without knowledge of boundary conditions but only with sparse trajectory measurements.

\begin{figure}[hbt!]
  \centering
  \includegraphics[width=0.5\textwidth]{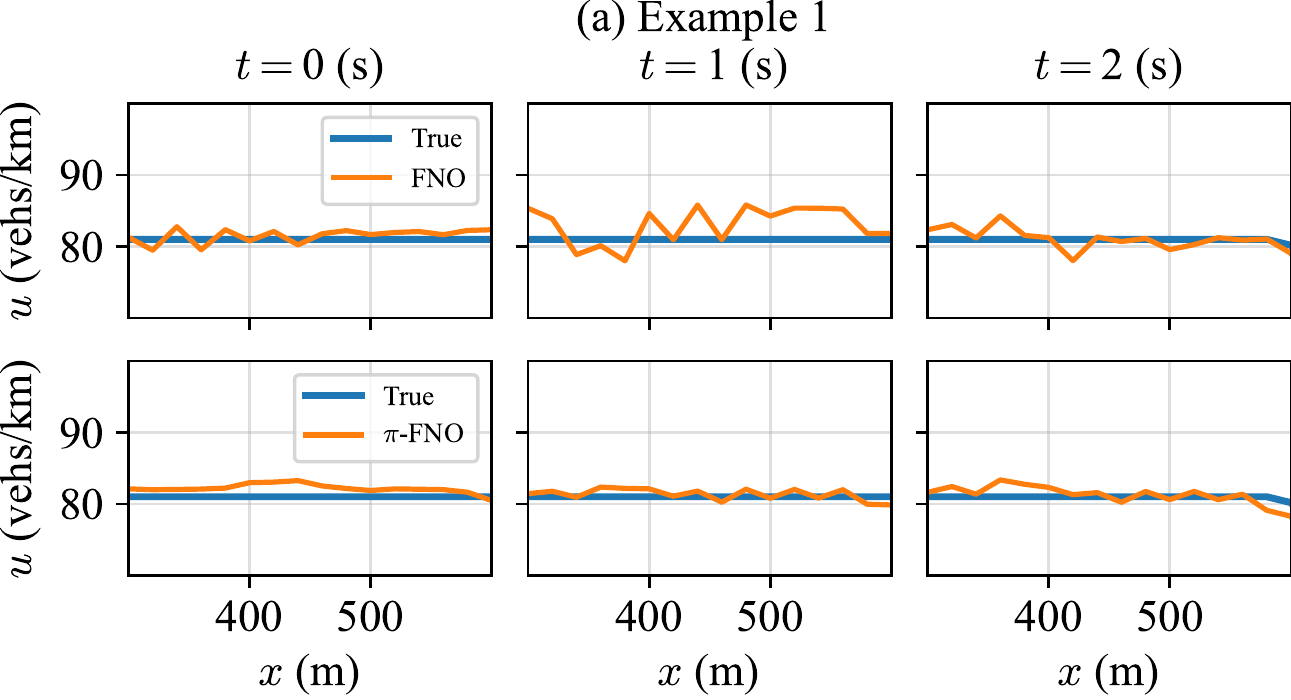} \hspace{0.04in}
  \includegraphics[width=0.5\textwidth]{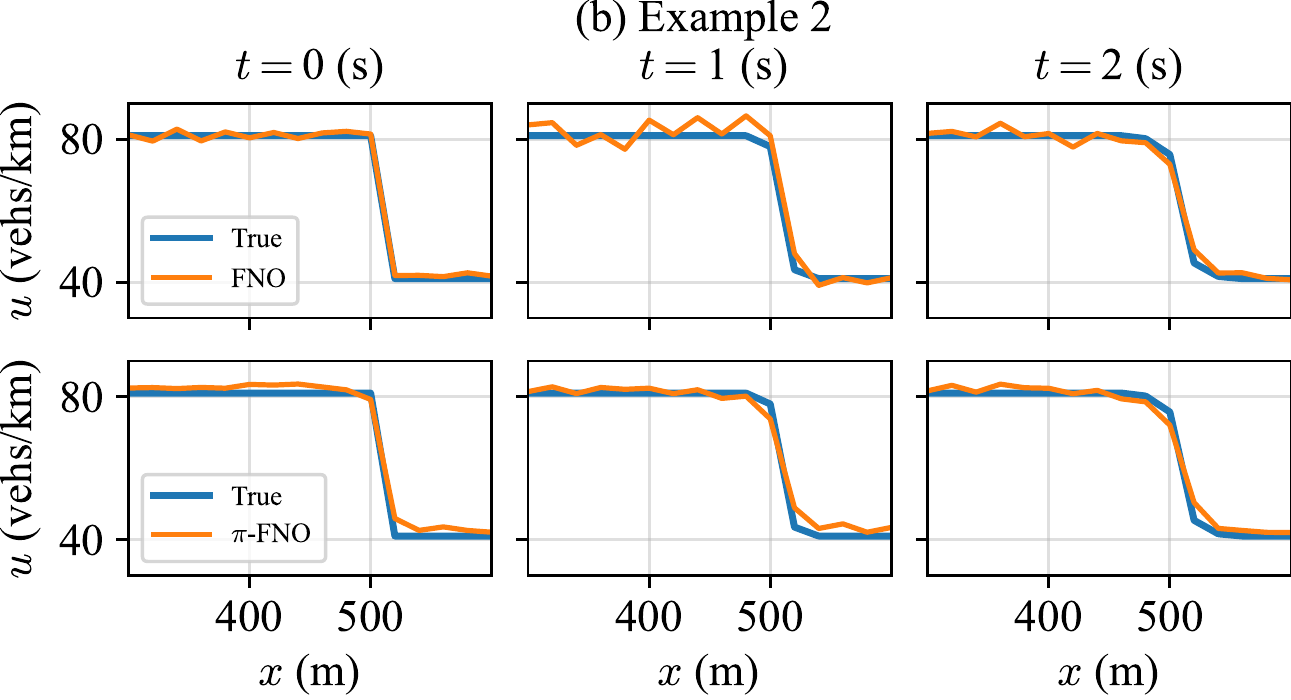}
  \caption{Solution profiles as a function of $x$ }
  \label{fig:shock_soln}
\end{figure}

\paragraph{Physics-informing on the behavior of shock solutions}
In Figure \ref{fig:shock_soln}, we compare solutions (zoomed-in) for a constant and a step-wise initial conditions. We see that the FNO model (physics-uninformed) produces noisy predictions, whereas the $\pi$-FNO model (physics-informed) smoothens these artifacts. This suggests the benefits of physics-informing in producing physically consistent solutions. 

\section{Summary and Discussion}
\label{sec:disc}

We explored the Fourier Neural Operator (FNO) in learning the weak solutions of scalar non-linear hyperbolic partial differential equations (H-PDEs), which has seen limited success in the deep learning-based computational literature. We focused on quantifying the generalization error of the FNO solver as a function of input complexity, taking vehicular traffic flow as an example. We found that the FNO solver can be trained using simple solutions and can easily generalize to complex inputs with an acceptable error tolerance $-$ the out-of-sample errors grew linearly with input complexity. We also showed the benefits of physics-informing in predicting physically consistent solutions, e.g., correct shock behavior. To the authors' knowledge, this is the first empirical study on generalization capabilities of learning-based solvers for non-linear H-PDEs.

A limitation of the current solver is that it requires a regular grid-like computational domain, partly due to the Fourier Transform operator. Our efforts continue to extend these solvers to irregular graph-like computational domains, e.g., to solve traffic flow on a city network.

\section{Broader Impact}

The modeling and control of dynamical systems such as road traffic, water supply, and communication networks are fundamental in functioning large-scale urban cities, which contribute to one-third of the global carbon footprint. The techniques developed in this study aid in building low-resource computational tools for controlling these dynamical systems, which are often modeled as partial differential equations. The data-driven learning paradigms studied in this work help advance the development of edge computing infrastructures such as connected vehicles, smart personal gadgets, and even augmented/virtual reality applications.


\section*{Acknowledgments}
This work was supported in part by the NYUAD Center for Interacting Urban Networks (CITIES), funded by Tamkeen under the NYUAD Research Institute Award CG001, and in part by the NYUAD Research Center on Stability, Instability, and Turbulence (SITE), funded by Tamkeen under the NYUAD Research Institute Award CG002. The views expressed in this article are those of the authors and do not reflect the opinions of CITIES, SITE, or their funding agencies.

\bibliographystyle{amsplain}
\bibliography{neurips_2022.bib}

\appendix

\begin{table*}[t!]
	\centering
	\caption{Hyper-parameters used in the study}
	\label{tab:hyperparams}
	\resizebox{\textwidth}{!}{%
		\begin{tabular}{@{}|ll|ll|ll|@{}}
			\toprule
			\multicolumn{2}{|l|}{\textbf{LWR simulation}} & \multicolumn{2}{l|}{\textbf{FNO model}} & \multicolumn{2}{l|}{\textbf{FNO training}} \\ \midrule
			\multicolumn{1}{|l|}{space dimension} & $1000$ m & \multicolumn{1}{l|}{\# Fourier layers $L$} & 4 & \multicolumn{1}{l|}{coefficient $\lambda$} & $2.0$ \\ \midrule
			\multicolumn{1}{|l|}{time dimension} & $600$ sec & \multicolumn{1}{l|}{\# modes in $x$ dimension} & 24 & \multicolumn{1}{l|}{\# epochs} & $500$ \\ \midrule
			\multicolumn{1}{|l|}{discretization size} & ($50 \times 600$) & \multicolumn{1}{l|}{\# modes in $t$ dimension} & 128 & \multicolumn{1}{l|}{batch size} & $128$ \\ \midrule
			\multicolumn{1}{|l|}{cell width $\Delta x$} & $20$ m & \multicolumn{1}{l|}{\# latent width} & 64 & \multicolumn{1}{l|}{learning rate} & $1e-3$ \\ \midrule
			\multicolumn{1}{|l|}{cell width $\Delta t$} & $1$ sec & \multicolumn{1}{l|}{\multirow{2}{*}{lifting operator  $\mathcal{P}$}} & \multirow{2}{*}{\begin{tabular}[c]{@{}l@{}}Linear layer\\ with depth 128\end{tabular}} & \multicolumn{1}{l|}{learning rate scheduler} & step-wise \\ \cmidrule(r){1-2} \cmidrule(l){5-6} 
			\multicolumn{1}{|l|}{max density $u_{\rm max}$} & $120$ vehs/km & \multicolumn{1}{l|}{} &  & \multicolumn{1}{l|}{optimizer} & Adam GD \\ \midrule
			\multicolumn{1}{|l|}{max flow $q_{\rm max}$} & $1800$ vehs/hr & \multicolumn{1}{l|}{\multirow{2}{*}{lifting operator $\mathcal{Q}$}} & \multirow{2}{*}{\begin{tabular}[c]{@{}l@{}}2-layer FNN \\ with depth 128\end{tabular}} & \multicolumn{1}{l|}{\# training samples} & $5200$ \\ \cmidrule(r){1-2} \cmidrule(l){5-6} 
			\multicolumn{1}{|l|}{} &  & \multicolumn{1}{l|}{} &  & \multicolumn{1}{l|}{\# testing samples} & $400$ \\ \bottomrule
		\end{tabular}%
	}
\end{table*}

\section{Godunov scheme for LWR simulation}
\label{app:godunov}

The LWR traffic flow model can be solved using Godunov's numerical scheme. 
Let $i$ and $j$ be the space and time index, denote $u_{(i,j)}$ as the average traffic density for each cell $(i,j)$. The traffic density is updated as:
\begin{equation} 
    u_{(i, j+1)} = u_{(i, j)} + \frac{\Delta t}{\Delta x} \left[ q_{(i-1/2, j)} - q_{(i+1/2, j)} \right]
\end{equation}
where $q_{(i-1/2, j)}$ is the cell boundary flux from cell $(i-1,j)$ to cell $(i,j)$. $\Delta t$ and $\Delta$ x denote the temporal and spatial width. The boundary flux is given by, 
\begin{equation*} 
    \begin{split}
        & q_{(i-1/2, j)} = \min \left\{ Q_{\rm dem}^{(i, j-1)}, ~Q_{\rm sup}^{(i, j)} \right\} \\
        & q_{(i+1/2, j)} = \min \left\{ Q_{\rm dem}^{(i, j)}, ~Q_{\rm sup}^{(i, j+1)} \right\}
        \end{split}
\end{equation*}
where 
\begin{equation*} 
    \begin{split}
        & Q_{\rm dem}^{(i, j)} = 
            \begin{cases}
                f \left( u^{(i, j)} \right) & \textrm{if}~~ u^{(i, j)} \leq u_{\rm cr} \\
                q_{\rm max} & \text{otherwise}
            \end{cases} \\
        & Q_{\rm sup}^{(i, j)} = 
            \begin{cases}
                f \left( u^{(i, j)} \right) & \textrm{if}~~ u^{(i, j)} > u_{\rm cr} \\
                q_{\rm max} & \text{otherwise}
            \end{cases}
        \end{split}
\end{equation*}
where $f(u)$ is the traffic flux function (also called Fundamental relation in traffic flow literature)
\[ 
f(u) = u (1-u/u_{max})v_{max}.
\]
where $v_{max}$ is maximum speed, $u_{max}$ is maximum density
 and $q_{\rm max}$ is the maximum traffic flux.
The above numerical scheme is run for different sets of initial and boundary conditions shown in Figure \ref{fig:inp_conds}. For the inverse problem, we draw random vehicle trajectories to represent $\bar{u}_b$. For all the testing, we used $10$ random vehicle trajectories as the input.

\section{Hyperparameters and training details}
\label{app:hyper}

The hyperparameters used in the study are summarized in Table \ref{tab:hyperparams}. The training hyperparameters are chosen by an independent trial-and-error experiment. We modified the original FNO implementation from \citet{li2021fno} to implement the $\pi$-FNO model in python using the PyTorch machine learning library; see Appendix \ref{app:data}. The training was performed on a GPU cluster (NVIDIA Tesla V100 32GB) and the total run time was around $\sim 45$ min for a single training experiment.

The objective function coefficient $\lambda$ is obtained by training a series of $\pi$-FNO models for $\lambda = \{0,0.05,0.1,\dots,0.95,1.0\}$. The $\lambda$ value corresponding to the least validation error is considered optimal.

\section{Datasets and codes}
\label{app:data}

The datasets, codes, and pretrained models are shared at \url{https://github.com/bilzinet/pifno} under the MIT license.

\end{document}